\documentclass[10pt,twocolumn,letterpaper]{article}

\usepackage{cvpr}              
\usepackage[dvipsnames]{xcolor}

\definecolor{cvprblue}{rgb}{0.21,0.49,0.74}
\usepackage[pagebackref,breaklinks,colorlinks,citecolor=cvprblue]{hyperref}
\usepackage{algorithm}
\usepackage{algpseudocode}
\usepackage{inconsolata}
\usepackage{microtype}

\usepackage{bm}
\newcommand{\spm}[1]{\tiny{$\,\pm$#1}}
\def\paperID{21} 
\def\confName{CVPR}
\def\confYear{2024}

\title{Consistency$\bm{^2}$: Consistent and Fast 3D Painting with Latent Consistency Models}

\author{Tianfu Wang \quad
Anton Obukhov \quad
Konrad Schindler
\vspace{5px}
\\
{\small Photogrammetry and Remote Sensing, ETH Zurich}
}

\begin{document}
\maketitle
\begin{abstract}
Generative 3D Painting is among the top productivity boosters in high-resolution 3D asset management and recycling.
Ever since text-to-image models became accessible for inference on consumer hardware, the performance of 3D Painting methods has consistently improved and is currently close to plateauing.
At the core of most such models lies denoising diffusion in the latent space, an inherently time-consuming iterative process.
Multiple techniques have been developed recently to accelerate generation and reduce sampling iterations by orders of magnitude. 
Designed for 2D generative imaging, these techniques do not come with recipes for lifting them into 3D.
In this paper, we address this shortcoming by proposing a Latent Consistency Model (LCM) adaptation for the task at hand.
We analyze the strengths and weaknesses of the proposed model and evaluate it quantitatively and qualitatively. 
Based on the Objaverse dataset samples study, our 3D painting method attains strong preference in all evaluations.
Source code is available at \href{https://github.com/kongdai123/consistency2}{https://github.com/kongdai123/consistency2}.
\end{abstract}    
\section{Introduction}
\label{sec:intro}
Recently, we witnessed a breakthrough in generative 3D content creation and painting~\cite{poole2022dreamfusion, richardson2023texture, chen2023fantasia3d}, empowering content creators and 3D artists to recycle old 3D assets or prototype texture design ideas using simple text prompts.
These methods either involve end-to-end optimization with the diffusion model as guidance through score distillation sampling (SDS)~\cite{poole2022dreamfusion}, which takes thousands of iterations to converge, or they iteratively denoise multi-view images in latent space, which involves tens to hundreds of denoising iterations for standard Latent Diffusion Models (LDMs)~\cite{rombach2022high}. 
On top of that, new techniques in diffusion sampling have been developed to speed up the generation process.
In particular, Latent Consistency Models (LCMs)~\cite{song2023consistency, luo2023latent} enable one-step generation by learning to directly map noise to data while retaining the option of multi-step sampling that trades runtime for better generation quality. 
Despite the significant speed-up that LCMs bring to image generation, exploration of their efficient operation in 3D has barely started.

\begin{figure*}[t]
    \centering
    \includegraphics[width=0.9\linewidth]{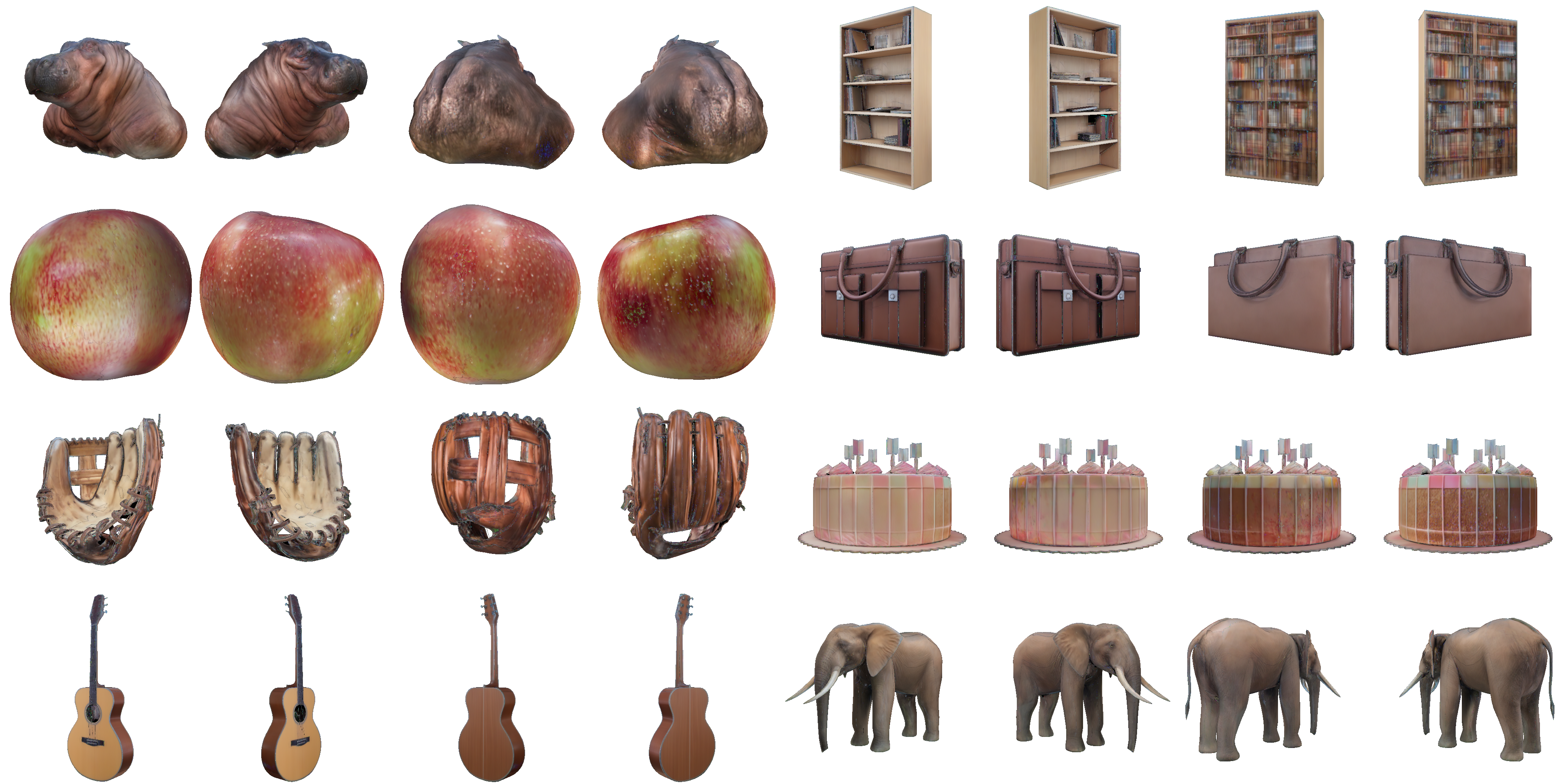}
    \caption{
    Selected painting results of Objaverse~\cite{deitke2023objaverse} meshes using Consistency$^2$. Our method paints detailed high-resolution textures with very few denoising diffusion steps and allows for free camera pose selection for view painting.
    }
    \label{fig:gallery}
\end{figure*}

We present \textit{Consistency${^2}$}, a fast way to generate multi-view \textit{consistent} surface textures of a mesh using latent \mbox{\textit{consistency}} models. 
Our work takes accelerated text-to-image diffusion models, particularly LCMs, and lifts their generative power to 3D in the task of generative mesh painting. 
The key to successful lifting lies in achieving view-consistent painting while not compromising speed and generation quality.
We tackle this problem from the perspective of \textit{multi-view denoising}, where we tailor our painting procedure to the few-step sampling process of LCMs. 
Our method can generate high-quality mesh paintings in 4 timesteps {per view}, taking less than two minutes {per mesh} on a single consumer GPU. 
\textit{Consistency${^2}$} provides a quick way for 3D content creators to prototype painting designs from text descriptions. 
Overall, it is much faster than most 
previous
mesh painting methods. 

\noindent
Our contributions can be summarised in the following:
\begin{itemize}

    \item We introduce the first recipe for mesh painting by multi-view denoising LCMs, benefiting from their few-step sampling process.
    
    \item We design a novel painting representation with separate noise and color textures. By applying appropriate interpolation techniques to each, we gain greater flexibility in texture resolution choice and camera view sampling.

    \item We apply our method to a diverse range of 3D meshes (Fig.~\ref{fig:gallery}), demonstrating competitive quality and runtime.
\end{itemize}
\section{Related Work}
\label{sec:rel}
\paragraph{Diffusion and Consistency Models}
Denoising diffusion probabilistic models (DDPMs)~\cite{ho2020denoising} have recently achieved state-of-the-art results in high-quality and photorealistic image generation.  
LDMs~\cite{rombach2022high, podell2023sdxl} shift the denoising process to low dimensional latent space, reducing computational requirements while achieving competitive quality.
Through frameworks such as ControlNet~\cite{zhang2023adding}, these models have been further conditioned on various modalities such as depth maps, images, and segmentation masks.

Numerous improvements have been proposed to accelerate image generation~\cite{song2020denoising, lu2022dpm, song2023consistency}. 
Consistency models, proposed by Song \etal~\cite{song2023consistency}, emerge as a new category of diffusion models that enable one- or few-step generation from noise to data. 
Consistency models are formulated with the property of \textit{self-consistency}, where points on the same Probability Flow ODE trajectory map to the same data sample~\cite{song2020score}.
LCMs~\cite{luo2023latent} extend consistency distillation to the latent space, allowing distillation from LDMs. 

\paragraph{Generating 3D with Diffusion Guided Optimization}
Alongside advancements in 2D image diffusion models, there has been an increasing interest in leveraging these models to generate 3D objects. 
One line of work is generation through end-to-end optimization of 3D scene parameters using a differentiable renderer, guided by the supervision of a 2D image diffusion model through Score Distillation Sampling (SDS)~\cite{poole2022dreamfusion, metzer2023latent, decatur20233d, chen2023fantasia3d, qian2023magic123, tang2023dreamgaussian, wang2024prolificdreamer, tsalicoglou2023textmesh}. 
This technique is first introduced in DreamFusion~\cite{poole2022dreamfusion} with Neural Radiance Fields (NeRF)~\cite{mildenhall2021nerf} as the 3D representation. 
Since then, this recipe has been applied to alternative 3D representations such as textured meshes~\cite{metzer2023latent, decatur20233d, tsalicoglou2023textmesh}, implicit surface representations~\cite{gao2022get3d, qian2023magic123}, BRDF materials~\cite{chen2023fantasia3d}, and 3D Gaussians~\cite{kerbl20233d, tang2023dreamgaussian}. 
Despite their flexibility, SDS methods converge slowly and produce oversaturated results~\cite{poole2022dreamfusion}.
Moreover, they do not benefit from LCM speedups as LCM is tailored to generation through \textit{denoising}.

\paragraph{Multi-view and Multi-patch Diffusion Generation}
Since diffusion models are often trained on images with a fixed resolution, there has been a lot of interest in using these models to generate surfaces of arbitrary size, such as panoramas~\cite{bar2023multidiffusion, lee2023syncdiffusion, wang2023generativepowers, zhang2023diffcollage} and 3D in the context of mesh painting~\cite{richardson2023texture, wang2023breathing, chen2023fantasia3d, cao2023texfusion}.
For panorama generation, MultiDiffusion~\cite{bar2023multidiffusion} proposes breaking down a large canvas into overlapping patches and fusing intermediate denoising results using a weighted average.
{Generative Powers-of-10~\cite{wang2023generativepowers} generates} view-consistent image zoom-ins by conducting multi-scale sampling across zoom levels in both color and noise spaces.
On the 3D side, a line of work exists on generating textures for a given 3D geometry using {\it multi-view denoising} with the same image diffusion model~\cite{richardson2023texture, wang2023breathing, chen2023text2tex, cao2023texfusion}.
TEXTure~\cite{richardson2023texture} pioneered this approach by denoising each camera view one after another and back-projecting each iteration to a texture map using rasterizer-based differentiable rendering. 
One work that shares the idea of combining the intermediate multi-view denoising results with our approach is TexFusion~\cite{cao2023texfusion}.
However, that method uses a single latent texture map, which limits its flexibility in terms of texture resolution and camera pose selection.

\section{Methods}
\label{sec:method}
\paragraph{Prerequisites}
Given a 3D mesh and a text description, we aim to achieve fast, view-consistent mesh painting.
We consider only pre-trained depth- and text-conditioned LCMs to ensure fast operation. 
Inspired by works such as TexFusion~\cite{cao2023texfusion} and TEXTure~\cite{richardson2023texture}, we parameterize the surface with 2D texture maps.
Compared to neural radiance or color fields exhibiting greater flexibility~\cite{wang2023breathing}, mesh textures are a desirable representation for painting since they are inherently view-consist (when mipmapped, see below) and come attached to the geometry.
Unlike TexFusion~\cite{cao2023texfusion} and \mbox{TEXTure}~\cite{richardson2023texture}, we employ two different texture map layers to render the noise state and the predicted content separately.
This design choice is inspired by Generative Powers-of-10~\cite{wang2023generativepowers}, where it is demonstrated that such a separation helps fuse a stack of 2D images at different zoom levels. 
We extrapolate this idea to meshes in 3D and consider the process of rendering a mesh with multiple virtual cameras similar to sampling overlapping patches on the texture map with different degrees of magnification, which depend on the local mesh curvature and distance to the camera.
However, Generative Powers-of-10~\cite{wang2023generativepowers} uses a proprietary diffusion model~\cite{qian2023magic123} operating directly in the RGB pixel space.
Contrary to that, modern high-quality open-source LDMs~\cite{rombach2022high} operate in the latent space to save computational resources.

We empirically found manipulating multi-view latents challenging and resulting in poor novel view synthesis. 
On the other hand, the RGB pixel space has well-established and understood strategies for interpolation, such as mipmapping~\cite{akenine2019real, Laine2020diffrast}.
Mipmapping involves an intermediate stack of downsampled textures. It automatically calculates the appropriate sampling level and the required texture pixels (texels) for interpolating each rendered pixel. 
With these observations in sight, the rationale for using an LCM as the generative model is twofold: 
(1) it enables fast and few-step generation (within ten timesteps), 
and (2) at each step, it can directly output a clean denoised latent sample, which can be immediately decoded into pixel space, and allows us to perform multi-view content fusion directly in the pixel space, taking advantage of mipmapping.

\paragraph{Variance Preserving Noise Rendering}
\begin{figure}[t]
    \centering
    \includegraphics[width=\linewidth]{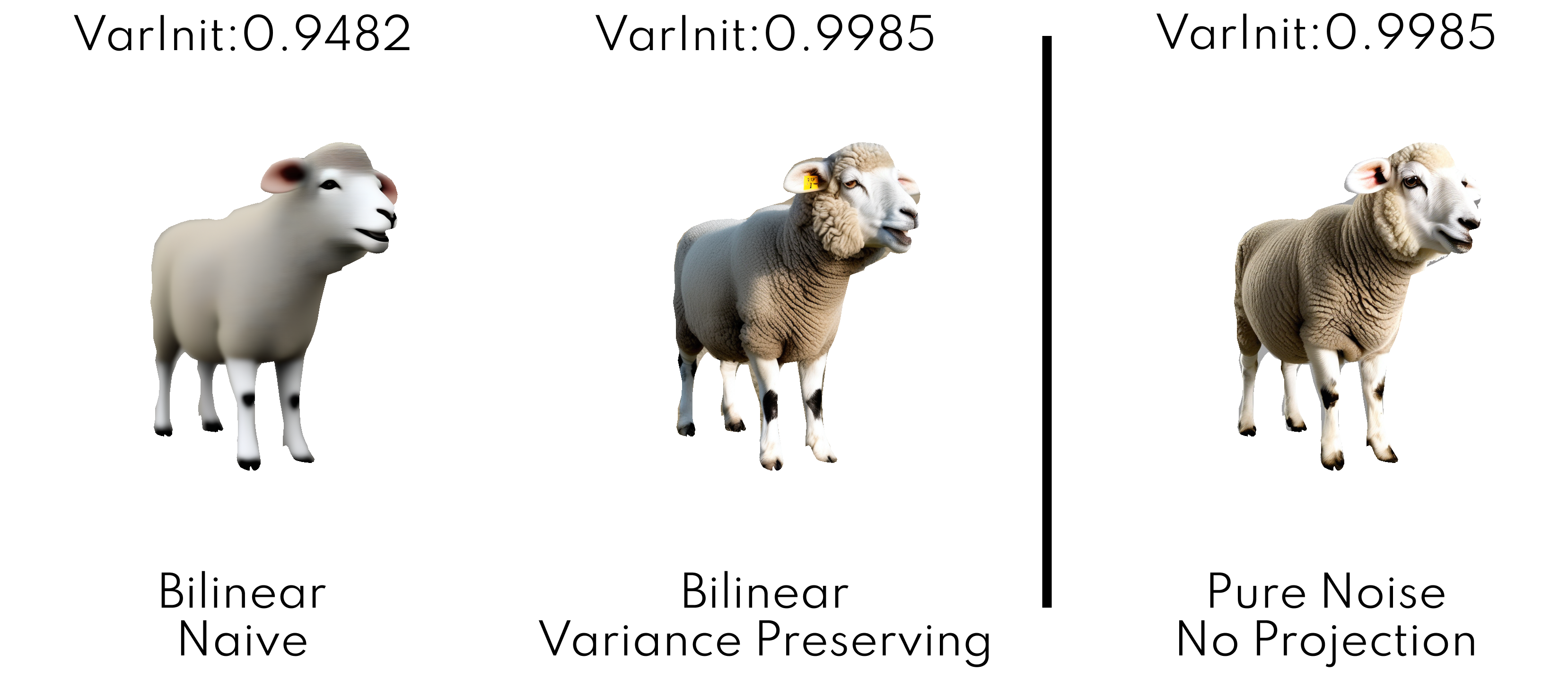}
    \caption{
    We project a mesh textured with latent noise using various texture interpolation methods.
    The naive bilinear interpolation (\textbf{left}) disturbs the latent probability distribution and gives poor results. 
    Our variance-preserving noise rendering (\textbf{middle}) results in a similar denoised quality of the mesh compared to just using a noise latent without any projection as a reference (\textbf{right}).  
    \texttt{VarInit} denotes the variance of the initial projected noise.
    }
    \label{fig:interp}
\end{figure}

Generative Powers-of-10~\cite{wang2023generativepowers} has shown that the image quality and consistency improve when noise is shared between patches. In the 3D painting case, we also want to render noise from multiple views to maximize sharing noise texels between views.
Compared to nearest neighbor interpolation, the bilinear interpolation of the textured surface facilitates better texel sharing across views.
However, the denoising diffusion process requires the projected noise component of the data to follow the normal distribution strictly.
As shown in Fig.~\ref{fig:interp}, bilinear interpolation leads to subpar results due to violating distribution properties.

We will now consider sampling from a random normal noise texture using bilinear interpolation.  
Specifically, the standard convex combination $Z' = rX + (1 - r)Y,$ of two i.i.d. random variables $X, Y \sim \mathcal{N}(0,1)$, where $r \! \in \! [0,1]$, is a random variable with non-unit variance: 
$
Z' \sim \mathcal{N}(0, r^2 + (1 - r)^2)
$
. 
The 2D case of bilinear interpolation involves 4 variables and two coefficients $r_u,r_v$, but inherently suffers from the same issue.
We designed a custom variance-preserving interpolation to alleviate this issue and ensure unit variance of the interpolated noise variable.
In the 1D case, it takes the following form:
\begin{equation}
\label{eq:interp}
    Z = \sqrt{r}X + \sqrt{(1 - r)}Y.
\end{equation}
Here $Z \!\sim\! \mathcal{N}(0,1)$ for all interpolation coefficients $r \!\in\! [0,1]$.
For 2D variance-preserving interpolation, we repeatedly apply the 1D case in Eq.~\eqref{eq:interp} in $u$ and $v$ dimensions. 

\paragraph{Multi-view Fusion Consistency Sampler}
\begin{figure}[b]
    \centering
    \includegraphics[width=0.8\linewidth]{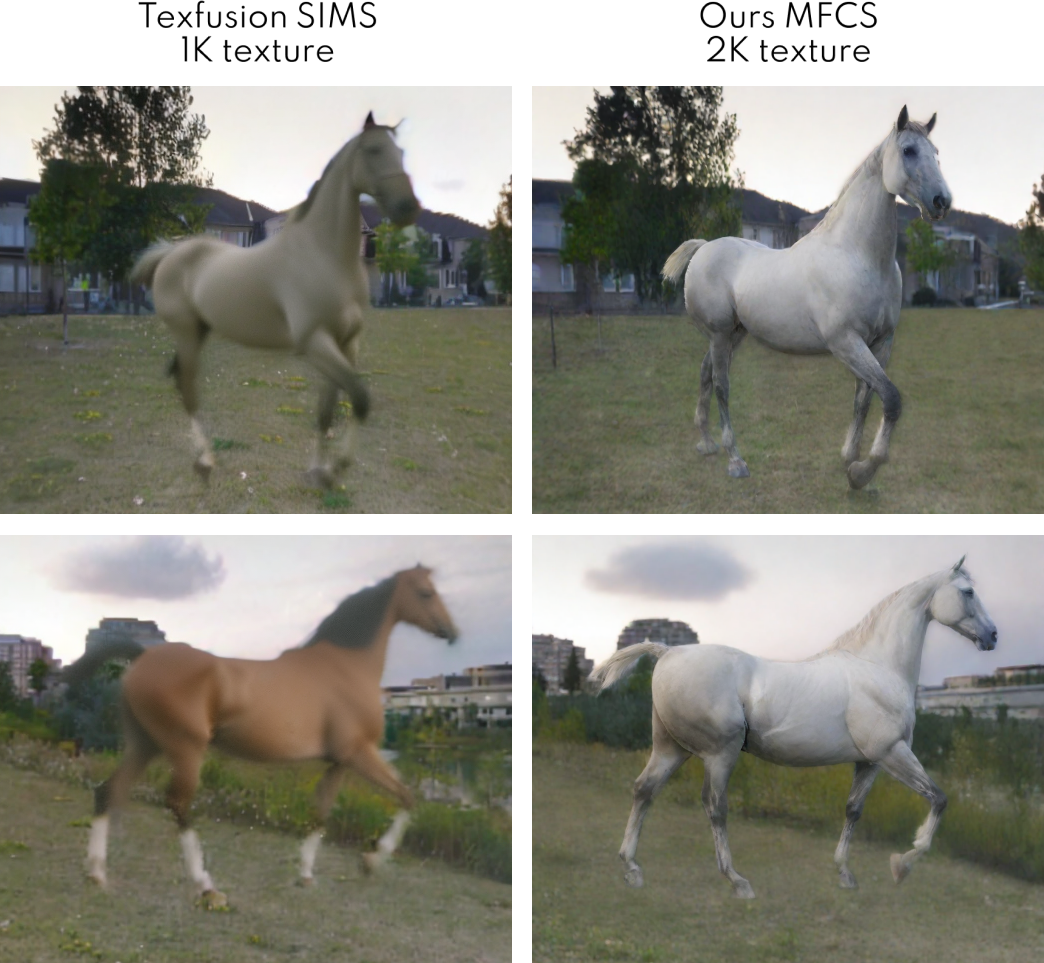}
    \caption{
    We show a comparison between Texfusion's Sequential Interlaced Multi-view Sampler (SIMS, \textbf{left})~\cite{cao2023texfusion} and our method (\textbf{right}) in accommodating high-resolution textures. 
    Texfusion's SIMS has rigid requirements on texture resolution and completely loses view consistency with a high-resolution texture. 
    Our formulation of separating noise and color textures enables high texture resolutions without sacrificing view consistency. 
    }
    \label{fig:const}
\end{figure}
We adapt the multi-view sampling and painting procedure to the sampling mechanism of LCMs~\cite{luo2023latent, song2023consistency}. 
First, the color texture is initialized to black, and the noise texture is initialized to random normal noise.
The object is rendered independently textured with color and noise for each camera view in each iteration.
Additionally, we store depth maps from the Z-buffer.
Color textures undergo mipmapping for better quality and artifact-free view consistency, and the background is filled from an environment cubemap. 
Noise texture rendering is performed with our custom variance-preserving interpolation, and the background is filled with random normal noise.
For each view $i$, we obtain the color latents $\mathbf{x}_i$ by converting the color-textured projections into the latent space with the LCM encoder.
We then compute a weighted average of the color ($\mathbf{x}_i$) and noise ($\mathbf{z}_i$) latents based on the LCM noise scheduling equation at timestep $t$: $\hat{\mathbf{x}}_{i, t} =  \alpha(t)\mathbf{x}_i +  \sigma(t)\mathbf{z}_i$. 
Refer to~\cite{luo2023latent} for the notation on $\alpha$ and $\sigma$.
The combined latent is then fed into the LCM, with the text prompts and depth map for conditioning. 
The resulting clean latent is produced in a single step and decoded back into RGB pixel space for multi-view fusion.
We perform the inverse rendering with mipmaps to update the color texture map.
Specifically, we optimize it to fit the denoised RGB images, minimizing the L2 photometric loss.
Due to the separation of noise and color textures and the usage of appropriate interpolation methods, we achieve flexibility in texture resolution, as shown in Fig.~\ref{fig:const}.
\section{Experiments}
\label{sec:exp}

\paragraph{Implementation and Setup}
We use the nvdiffrast~\cite{Laine2020diffrast} renderer to implement our variance-preserving interpolation. 
Additionally, PyTorch3D~\cite{ravi2020accelerating} framework is used for mesh and view processing. 
For the image generation model, we use SDXL~\cite{podell2023sdxl} model distilled to an LCM, with additional depth conditioning from ControlNet~\cite{zhang2023adding}. 
A dome of $15$ cameras covers the whole mesh located in the origin. 
For each iteration, we adjust the azimuth of each camera pose by $10^\circ$ to promote coverage of the object from various view angles. 
Additionally, we add relative pose descriptions in the prompt for each view to facilitate view consistency.
In our Multi-view Fusion Consistency Sampler, we use 4 denoising iterations, with text guidance scale $7.5$ and depth guidance scale $0.4$ on all iterations. 
The color texture resolution is $4K$, and the noise texture resolution is $768\times768$. 

\paragraph{Comparison to State-of-the-Art} 
We compare our method with Text2Tex~\cite{chen2023text2tex}, one of the best open-source state-of-the-art mesh painting methods. 
Each method is evaluated using a subset of the Objaverse evaluation meshes~\cite{deitke2023objaverse} from Text2Tex, totaling $343$ meshes in $203$ categories. 
We include text descriptions of meshes provided by Objaverse in the prompt.

\begin{figure}[t]
    \centering
    \includegraphics[width=0.8\linewidth]{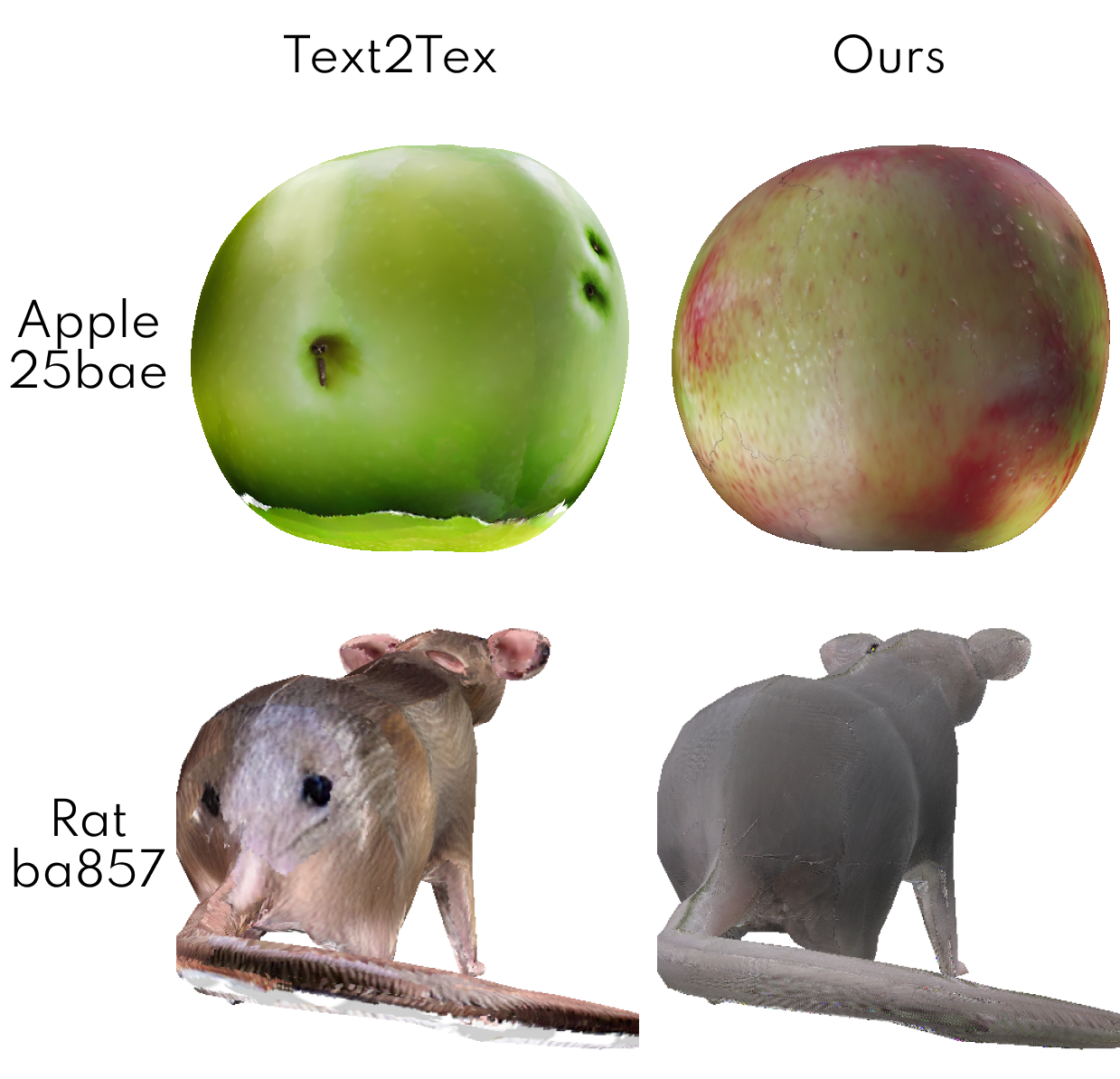}
    \caption{
    \textbf{Qualitative comparison of Objaverse~\cite{deitke2023objaverse} mesh painting.} 
    Text2Tex's~\cite{chen2023text2tex} sequential nature makes it susceptible to irrecoverable artifacts such as seams and coherence issues. 
    Our method is free of these limitations.
    }
    \label{fig:comp}
\end{figure}

\begin{table}[t]
\centering
\resizebox{0.8\linewidth}{!}{
\begin{tabular}{lcc}

\toprule

Method      & FID~\cite{heusel2017gans}~$\downarrow$ & KID ($\times$10\textsuperscript{-3})~\cite{binkowski2018demystifying}~$\downarrow$ \\ 

\midrule

Text2Tex \cite{chen2023text2tex}  & 28.93      & 6.88  \spm 0.05 \\

Consistency${^2}$ (Ours)      & \textbf{22.74}      & \textbf{4.02} \spm  0.03  \\ 

\bottomrule

\end{tabular}
}
\caption{
\textbf{Quantitative comparison of mesh painting methods.}
Generative metrics for Text2Tex~\cite{chen2023text2tex} and our method demonstrate superior painting of Objaverse~\cite{deitke2023objaverse} meshes.
}
\label{tab:metrics}
\end{table}

We sample views of the painted meshes from a set of object-centric views with various elevation angles. 
We also generate reference images using SDXL~\cite{podell2023sdxl} pipeline with $50$ denoising iterations guided by captions and depth maps obtained from the same views. 
This set of view-inconsistent images serves as a reference for the best generation quality attainable with a pure diffusion model.

We calculated~\cite{obukhov2020torchfidelity} Frechet Inception Distance (FID)~\cite{heusel2017gans} and Kernel Inception Distance (KID)~\cite{binkowski2018demystifying} between sets of painted and reference images. 
Our method outperforms Text2Tex in both metrics (Tab.~\ref{tab:metrics}).
We also recorded ${\sim}7.5\times$ runtime speed-up of our method (under $2$ minutes per mesh) compared to Text2Tex (around $15$ minutes per mesh)~\cite{chen2023text2tex}.

In a qualitative comparison, our approach of denoising views simultaneously with an LCM demonstrates better visual quality than sequential denoising methods such as Text2Tex.
Specifically, our method is free of seams and the Janus (multi-head) problem  (Fig.~\ref{fig:comp}).
\section{Conclusion}
\label{sec:discuss}
We presented a novel recipe for lifting LCM into 3D for fast generative painting. 
We introduced a creative mesh texture formulation, which separates noise and color and applies correct interpolation to both modalities, achieving unconstrained camera view selection and texture resolution. 
We conducted a study comparing our method with a state-of-the-art mesh painting method, Text2Tex, and showed advantages in generation quality and runtime. 
We believe our work is a step towards interactive and fast 3D painting that can benefit designers and content creators. 
\small
\bibliographystyle{ieeenat_fullname}
\bibliography{main}

\end{document}